\documentclass[10pt,twocolumn,letterpaper]{article}

\usepackage[pagenumbers]{cvpr}

\usepackage{graphicx}
\usepackage{amsmath}
\usepackage{amssymb}
\usepackage{booktabs}
\usepackage[dvipsnames]{xcolor}

\usepackage[pagebackref,breaklinks,colorlinks]{hyperref}

\usepackage[capitalize]{cleveref}
\crefname{section}{Sec.}{Secs.}
\Crefname{section}{Section}{Sections}
\Crefname{table}{Table}{Tables}
\crefname{table}{Tab.}{Tabs.}

\def\cvprPaperID{53} 
\def\confName{CVPR}
\def\confYear{2023}

\begin{document}

\title{Knowledge Assembly: Semi-Supervised Multi-Task Learning from Multiple Datasets with Disjoint Labels}

\author{Federica Spinola, Philipp Benz, Minhyeong Yu, Tae-hoon Kim\\
Deeping Source Inc., Seoul, Republic of Korea\\
{\tt\small \{federica.spinola, philipp.benz, minhyeong.yu, pete.kim\}@deepingsource.io}
}
\maketitle

\begin{abstract}
In real-world scenarios we often need to perform multiple tasks simultaneously. Multi-Task Learning (MTL) is an adequate method to do so, but usually requires datasets labeled for all tasks. We propose a method that can leverage datasets labeled for only some of the tasks in the MTL framework. Our work, Knowledge Assembly (KA), learns multiple tasks from disjoint datasets by leveraging the unlabeled data in a semi-supervised manner, using model augmentation for pseudo-supervision. Whilst KA can be implemented on any existing MTL networks, we test our method on jointly learning person re-identification (reID) and pedestrian attribute recognition (PAR). We surpass the single task fully-supervised performance by $4.2\%$ points for reID and $0.9\%$ points for PAR.
\end{abstract}

\section{Introduction}
\label{sec:intro}
With deep learning making its way into real products, it is crucial to deploy models that can efficiently and effectively understand the world around us. In scenarios like autonomous driving or surveillance, multiple tasks need to be solved simultaneously. Whilst networks can be trained for each individual task, this can lead to large inference times. In contrast, multi-task learning (MTL) aims to share information between tasks by learning them jointly, usually within a single network. Thus it can improve the individual tasks' performance \cite{zhang2021survey} and overall inference time. MTL has been successfully applied to many computer vision tasks \cite{kokkinos2017ubernet}. Nevertheless, it usually considers a single dataset labeled for all tasks \cite{cordts2016city, liu2015deep}, whereas in real-world applications we are often faced with a scarcity of labeled data for the tasks at hand: each dataset is only labeled for one of the tasks. Moreover, the human effort required to label all the datasets with the missing annotations would be time-consuming and impractical. Therefore, there is a need to tackle this issue and leverage partially labeled datasets in a MTL framework. 

\begin{figure}[t]
  \centering
   \includegraphics[width=0.9\linewidth]{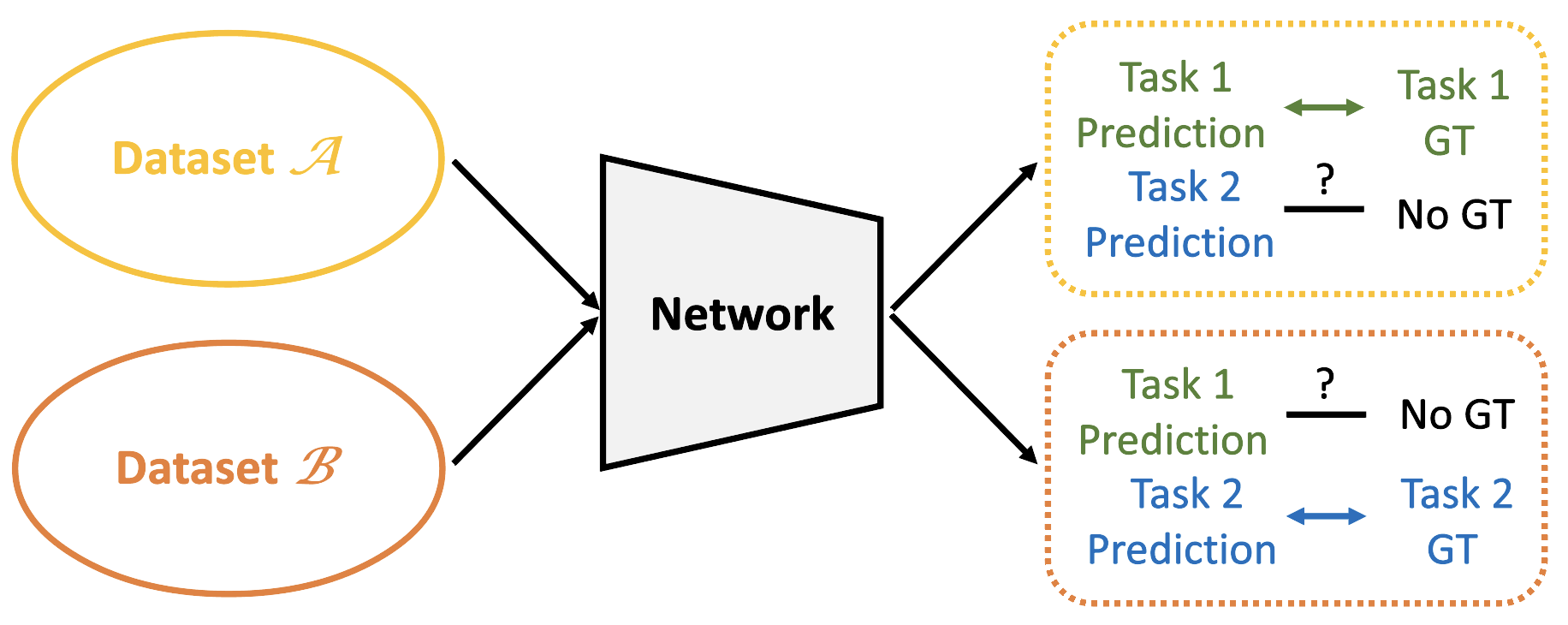}
   \caption{\textbf{Problem Definition.} The aim is to jointly learn two tasks, $1$ and $2$, given two datasets, $\mathcal{A}$ and $\mathcal{B}$ that only have labels for one of the tasks: dataset $\mathcal{A}$ for task $1$ and dataset $\mathcal{B}$ for task $2$.}
   \vspace{-0.4em}
   \label{fig:pb def}
\end{figure}

\begin{figure*}[ht]
  \centering
   \includegraphics[width=0.9\linewidth]{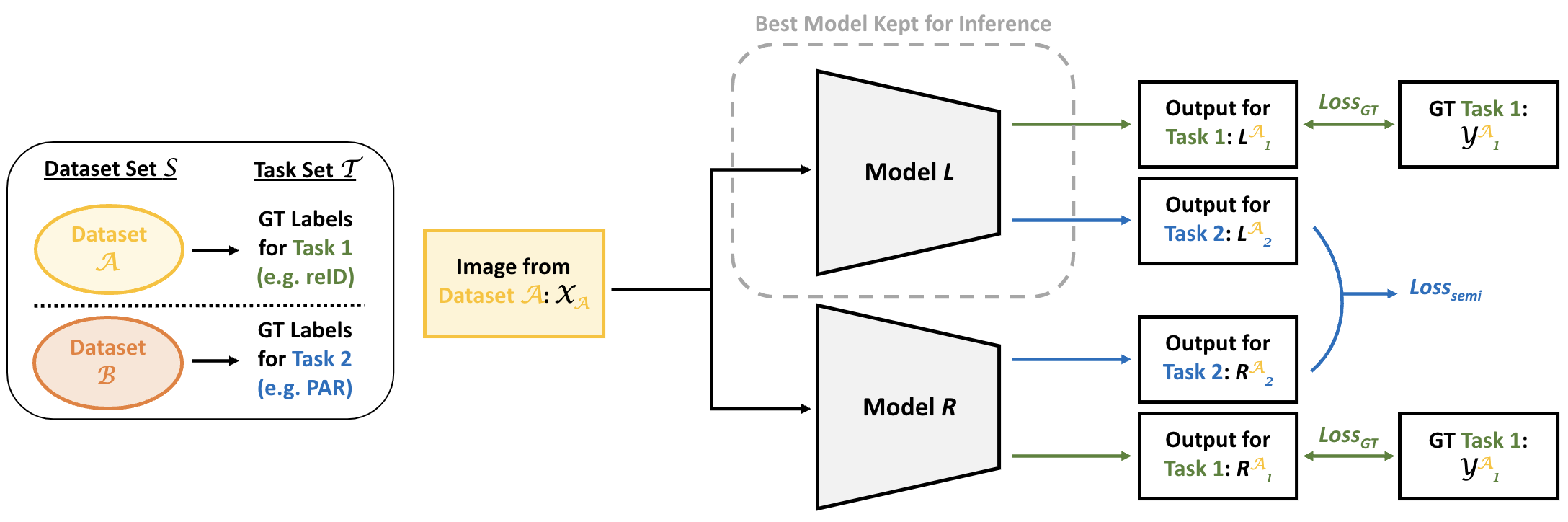}
   \caption{\textbf{KA Method Overview.} For training we initialize two instances $L$ and $R$ of the same model. A mini-batch containing both datasets $\mathcal{A}$ and $\mathcal{B}$ is fed into the networks, and outputs for both tasks $\mathcal{T}_1$ and $\mathcal{T}_2$ are predicted. If the corresponding input has a ground-truth (GT) label then a supervised loss $\mathcal{L}^{GT}$ is computed, else a semi-supervised loss $\mathcal{L}^{semi}$ is calculated. The latter loss ensures the outputs' consistency between the two network. In this Figure we show the forward path for one instance of dataset $\mathcal{A}$ (in \textcolor{Goldenrod}{yellow}) that only has GT labels for task $\mathcal{T}_1$ (in \textcolor{ForestGreen}{green}). During inference, the best of the two models is kept. Best viewed in color.}
   \label{fig:onecol}
\end{figure*}
In this work we consider the problem of learning multiple tasks from disjoint datasets, such that, for any given input image, the supervisory information is available for at least one but not all tasks at hand. An illustration of our problem setting is shown in Figure \ref{fig:pb def}. In the considered setting the training data can come from different domains, the domains can be represented in an unbalanced way, and supervision signals can be sparse. We could devise several methods to compute only supervised losses when ground-truth labels are available. However, we would discard a lot of information from the images that could otherwise be leveraged to guide the learning of the other tasks. Semi-supervised learning (SSL) demonstrated that using additional unlabeled data can improve the network's performance and enhance its generalization capabilities \cite{zhu2005semi}. Therefore, we propose Knowledge Assembly (KA), which aims to use all the data available within a MTL framework by leveraging SSL techniques.  \\
Our contributions are the following:
\vspace{-0.3em}
\begin{itemize}
    \setlength\itemsep{-0.3em}
    \item We propose a novel framework to train a single model for MTL by leveraging disjoint datasets.
    \item We suggest to utilize the unlabeled data for each task by introducing a semi-supervised consistency loss resulting from model augmentation.
    \item We evaluate our framework on the person re-identification (reID) and pedestrian attribute recognition (PAR) tasks, and show that our method outperforms all baselines.
\end{itemize}
\section{Related Work}
\label{sec:related work}
\label{sec:method}

\noindent
\textbf{Multi-Task Learning.}
The MTL framework is being successfully applied to multiple computer vision domains, especially in the realm of scene understanding \cite{vandenhende2021multi, zhen2020joint, kendall2018multi}. MTL is also used in the domain of face analysis \cite{huang2021age} or person identification \cite{schumann2017person, lin2019improving}. Most MTL techniques assume a single dataset labeled for all the relevant tasks, or use knowledge transfer across tasks (to learn more about MTL please refer to \cite{ruder2017overview}). However, they do not tackle the simultaneous learning of multiple tasks from different datasets with partial labels. 

\vspace{0.25cm}
\noindent
\textbf{Semi-Supervised Learning.}
In deep learning, the need for quality labeled data is crucial. However, more often than not, we encounter tasks for which labeled data is hard to get and cumbersome to annotate. SSL is needed to train models with partially unlabeled data \cite{ouali2020overview}. Successful methods for semi-supervised object classification \cite{xie2020unsupervised, sohn2020fixmatch} consist of using a self-consistency loss to ensure that a network outputs similar predictions for different input image augmentations. Conversely, \cite{chen2021semi} develops a semi-supervised framework based on model augmentation for semi-supervised semantic segmentation. Instead of using weak augmented images as pseudo-labels for strong augmented images, \cite{chen2021semi} initializes two instances of the same network and uses the output from one network to supervise the other network (and vice-versa). However, semi-supervised learning is usually restricted to scenarios where the labeled and unlabeled data come from the same distribution, and few works successfully manage to leverage semi-supervised learning in cross-domain settings \cite{gu2022contrastive}. Moreover, the above-mentioned works do not tackle the problem of semi-supervised learning in a MTL framework.

\vspace{0.25cm}
\noindent
\textbf{Multiple Datasets with Disjoint Labels.}
A few works tackle the problem of learning multiple tasks from disjoint datasets \cite{hong2020beyond, huang2020partly, dhar2022eyepad++, kim2018disjoint, wang2022semi}. The general approach is to use a MTL framework and deal with the encountered forgetting effect by leveraging knowledge distillation \cite{hong2020beyond, dhar2022eyepad++, kim2018disjoint}. However, these methods do not leverage the unlabeled data from one of the tasks to improve the other task's performance. Thus, other works \cite{huang2020partly, wang2022semi} use adversarial learning jointly with SSL to learn multiple tasks from disjoint datasets, whilst utilizing all the data at hand for each task. In our work, we also focus on the problem of MTL from disjoint datasets, and leverage unlabeled data using semi-supervised learning. Moreover, our training strategy can directly be applied to existing networks and does not require additional adversarial network designs. 

\section{Method}
In this Section, we will present our method, which is illustrated in Figure \ref{fig:onecol}.

\subsection{Problem Definition}
Following a similar setup to that of \cite{wang2022semi}, we consider a set of tasks $\mathcal{T}$, an input space consisting of a set of partially labeled datasets $\{\mathcal{X}_s\}_{s\in\mathcal{S}}$ and an output space consisting of a set of task-specific outputs $\{\mathcal{Y}_t\}_{t\in\mathcal{T}}$. $\mathcal{S}$ is the set of all datasets, where each dataset has labels for at least one of the tasks but can have missing labels for all the other tasks. We wish to train a single model $\mathcal{F}(x;\Theta,\theta_t): {\mathcal{X}_s}\rightarrow{\mathcal{Y}_t}\ \forall s\in\mathcal{S}, t\in\mathcal{T}$, where $\Theta$ are the shared network parameters and $\theta_t$ are the task specific network parameters. In this work we consider two input datasets ($\mathcal{S}_{\mathcal{A}}$ and $\mathcal{S}_{\mathcal{B}}$) and two tasks ($\mathcal{T}_{1}$ and $\mathcal{T}_{2}$), where dataset $\mathcal{S}_{\mathcal{A}}$ has only ground-truth (GT) labels $\mathcal{Y}^{\mathcal{A}}_1$ for task $\mathcal{T}_{1}$ and $\mathcal{S}_{\mathcal{B}}$ has only GT labels $\mathcal{Y}^{\mathcal{B}}_2$ for $\mathcal{T}_{2}$. 

\subsection{Model Augmentation}
Our method is designed within a SSL framework similar to the one proposed by \cite{chen2021semi}: consistency regularization through network perturbation. Indeed, consistency regularization has been proven to be an effective learning signal for SSL as it forces outputs to be consistent under different perturbations \cite{xie2020unsupervised, sohn2020fixmatch, chen2021semi}. We initialize two instances of the same network, $\mathcal{F}_{L}(x;\Theta^L)$ and $\mathcal{F}_R(x;\Theta^R)$, and feed any input image through both networks. The semi-supervised consistency loss is designed as follows: the output $L$ of model $\mathcal{F}_L$ is used as a pseudo-label to supervise the output $R$ of model $\mathcal{F}_R$, and vice-versa. 

More specifically, we include this semi-supervised training strategy within a MTL framework. Thus for an input $\mathcal{X}^{\mathcal{A}}$ ($\mathcal{X}^{\mathcal{B}}$) there will be two outputs for each task (four in total): $L^{\mathcal{A}}_1$ ($L^{\mathcal{B}}_1$) and $R^{\mathcal{A}}_1$ ($R^{\mathcal{B}}_1$) for task $\mathcal{T}_{1}$ (green path in Figure \ref{fig:onecol}) from models $\mathcal{F}_L$ and $\mathcal{F}_R$, and $L^{\mathcal{A}}_2$ ($L^{\mathcal{B}}_2$) and $R^{\mathcal{A}}_2$ ($R^{\mathcal{B}}_2$) for task $\mathcal{T}_{2}$ (blue path in Figure \ref{fig:onecol}) from models $\mathcal{F}_L$ and $\mathcal{F}_R$. Input $\mathcal{X}^{\mathcal{A}}$ ($\mathcal{X}^{\mathcal{B}}$) has an associated GT label $\mathcal{Y}^{\mathcal{A}}_1$ ($\mathcal{Y}^{\mathcal{B}}_2$) for task $\mathcal{T}_{1}$ ($\mathcal{T}_{2}$), with which a supervised loss $\mathcal{L}^{GT}_1$ ($\mathcal{L}^{GT}_2$) can be computed for both models. However, $\mathcal{X}^{\mathcal{A}}$ ($\mathcal{X}^{\mathcal{B}}$) does not have any GT label for task $\mathcal{T}_{2}$ ($\mathcal{T}_{1}$). Therefore, a semi-supervised consistency loss $\mathcal{L}_2^{semi}$ ($\mathcal{L}_1^{semi}$) is computed by using the output $L^{\mathcal{A}}_2$ ($L^{\mathcal{B}}_1$) as pseudo-supervision for $R^{\mathcal{A}}_2$ ($R^{\mathcal{B}}_1$), and by using the output $R^{\mathcal{A}}_2$ ($R^{\mathcal{B}}_1$) as pseudo-supervision for $L^{\mathcal{A}}_2$ ($L^{\mathcal{B}}_1$). 

\subsection{Objective function}
The networks are trained by computing losses over mini-batches. Each mini-batch is created by sampling data from both datasets: samples $\mathcal{S}^{b}_{\mathcal{A}}$ from dataset $\mathcal{S}_{\mathcal{A}}$ and samples $\mathcal{S}^{b}_{\mathcal{B}}$ from dataset $\mathcal{S}_{\mathcal{B}}$. We compute all the relevant task-specific supervised and semi-supervised losses.  
The total supervised loss $\mathcal{L}^{GT}$ is calculated as:

\begin{align}
    & \mathcal{L}^{GT} = \frac{1}{|\mathcal{S}^{b}_{\mathcal{A}}|}\sum_{s\in\mathcal{S}^{b}_{\mathcal{A}}} (\mathcal{L}^{GT}_1(L^{\mathcal{A}}_1, \mathcal{Y}^{\mathcal{A}}_1) + \mathcal{L}^{GT}_1(R^{\mathcal{A}}_1, \mathcal{Y}^{\mathcal{A}}_1)) \nonumber \\
    & + \frac{1}{|\mathcal{S}^b_{\mathcal{B}}|}\sum_{s\in\mathcal{S}^b_{\mathcal{B}}} (\mathcal{L}^{GT}_2(L^{\mathcal{B}}_2, \mathcal{Y}^{\mathcal{B}}_2) + \mathcal{L}^{GT}_2(R^{\mathcal{B}}_2, \mathcal{Y}^{\mathcal{B}}_2))
\end{align}

The consistency loss on unlabeled data $\mathcal{L}^{semi}_u$ is computed as: 

\begin{align}
    & \mathcal{L}^{semi}_u = \frac{1}{|\mathcal{S}^{b}_{\mathcal{A}}|}\sum_{s\in\mathcal{S}^{b}_{\mathcal{A}}}
    (\mathcal{L}^{semi}_2(L^{\mathcal{A}}_2, R^{\mathcal{A}}_2) + \mathcal{L}^{semi}_2(R^{\mathcal{A}}_2, L^{\mathcal{A}}_2)) \nonumber \\
    & + \frac{1}{|\mathcal{S}^{b}_{\mathcal{B}}|}\sum_{s\in\mathcal{S}^{b}_{\mathcal{B}}}
    (\mathcal{L}^{semi}_1(L^{\mathcal{B}}_1, R^{\mathcal{B}}_1) + \mathcal{L}^{semi}_1(R^{\mathcal{B}}_1, L^{\mathcal{B}}_1))
\end{align}

We also compute the consistency loss for labeled data $\mathcal{L}^{semi}_l$. Therefore the total semi-supervised loss becomes: $\mathcal{L}^{semi} = \mathcal{L}^{semi}_u + \mathcal{L}^{semi}_l$. \\
Finally, the total training objective for each batch is the following:
\begin{equation}
    \mathcal{L} = \mathcal{L}^{GT} + \lambda \mathcal{L}^{semi}
    \label{eq: tot loss}
\end{equation}
Where $\lambda$ is a hyperparameter to balance the supervised and consistency losses.

\begin{table*}[t]
\centering
\caption{\textbf{Quantitative Experiment Results on ResNet-18.} We train our method on Market1501 and PA100K and evaluate reID and PAR on the test-sets of Market1501 and PA100K respectively. Best results are shown in \textbf{bold}.}
\label{tab:res same domain}
\scalebox{0.8}{
\begin{tabular}{l|cccc|cccc}
\hline
{\textit{ResNet-18}} & \multicolumn{4}{c|}{Market1501 (reID)} & \multicolumn{4}{c}{PA100K (PA)} \\ \cline{2-9} 
& \multicolumn{1}{c}{mAP (\%)} & \multicolumn{1}{c}{rank-1 (\%)} & \multicolumn{1}{c}{rank-5 (\%)} & \multicolumn{1}{c|}{rank-10 (\%)} & \multicolumn{1}{c}{ma (\%)} & \multicolumn{1}{c}{F1 (\%)} & \multicolumn{1}{c}{Prec. (\%)} & \multicolumn{1}{c}{Rec. (\%)} \\ \hline
reID only  & 52.6 & 72.2 & 88.2 & 92.2 & -    & -    & -      & -      \\
PAR only   & -    & -    & -    & -    & 74.3 & 82.3 &  84.4  &   80.3 \\
reID ssl   & 48.9 & 69.6 & 86.2 & 91.1 & -    & -     &    -     &-   \\
PAR ssl    &  -    &  -  &  -   &  -   & \bf{75.1} & 83.1 & 85.0 & \bf{81.3} \\
MTL baseline & 56.4 & 75.1 & \bf{90.1} & \bf{93.7} & 74.3 & 82.3 & 84.4 & 80.3 \\
KA (ours)  & \bf{56.8} & \bf{75.9} & 89.8 & \bf{93.7} & 74.4 & \bf{83.2} & \bf{85.2}   & \bf{81.3} \\ \bottomrule
\end{tabular}
}
\end{table*}

\begin{figure}[t]
  \centering
   \includegraphics[width=0.8\linewidth]{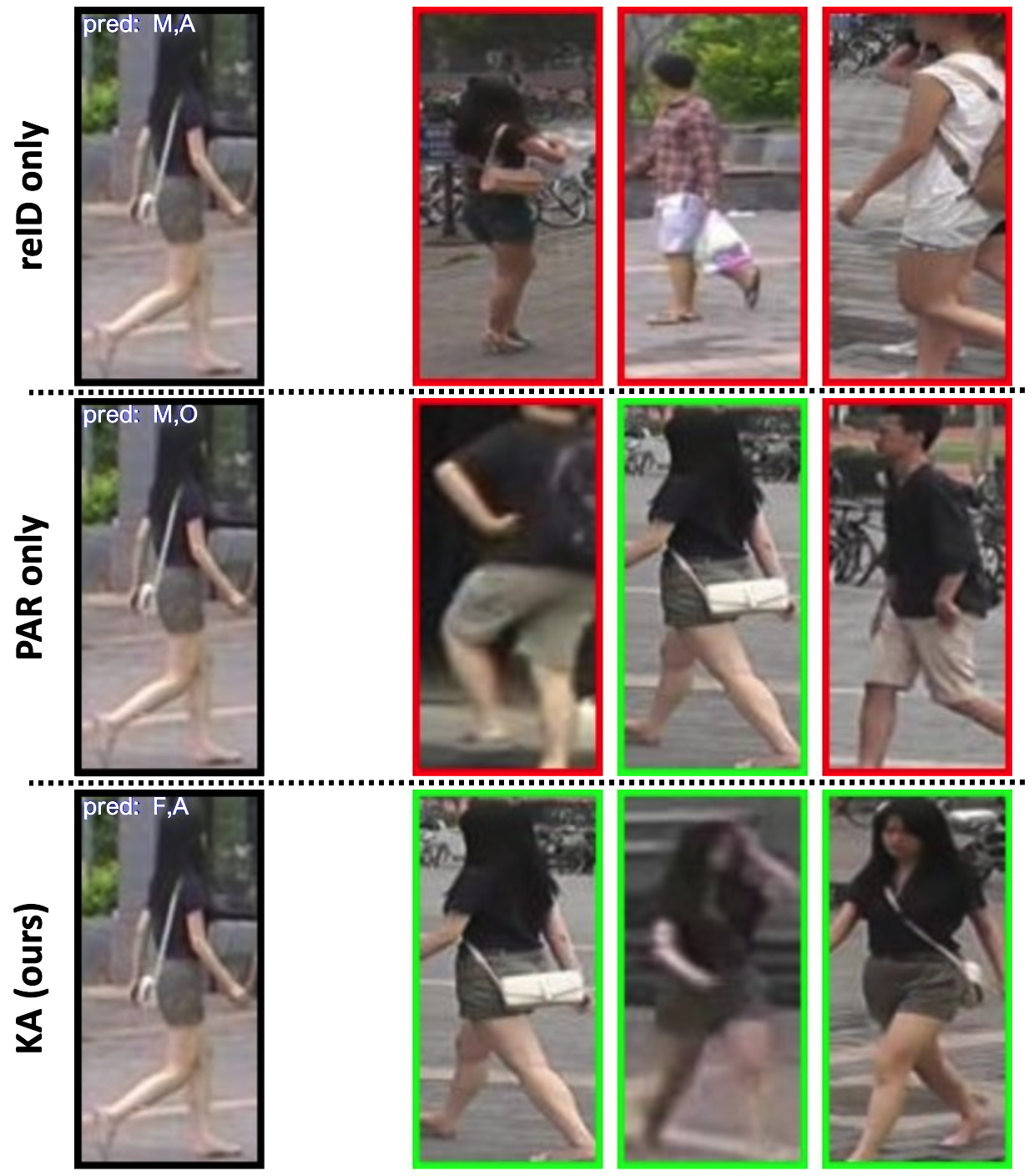}
   \caption{\textbf{Qualitative Results on Market1501.} We show the top three reID results (colored on the right) for a query image (black on the left). Correct reID is in \textcolor{green}{green} and wrong reID is in \textcolor{red}{red}. The query's attributes are also predicted. We display gender (M: male, F: female) and age (Y: young, A: adult, O: old). Best viewed in color.}
   \vspace{-0.3em}
   \label{fig:qualita}
\end{figure}

\section{Experiments}
\label{sec:results}
Whilst our method can be applied to any two or more related tasks, we implement it to the tasks of person reID and PAR ($\mathcal{T}_{1}$ corresponds to reID and $\mathcal{T}_{2}$ to PAR). The experimental details and results are outlined in Section \ref{sec:results} below. 
\subsection{Datasets and Metrics}
For all our experiments, we train our KA on the Market1501 \cite{zheng2015scalable} (contains only reID GT labels) and the PA100K \cite{liu2017hydraplus} (contains only PAR GT labels from the UPAR dataset \cite{specker2023upar}) training datasets. We then evaluate reID on the test set of Market1501 and evaluate PAR on the test set of PA100K. 

For person reID, we evaluate our performance based on the mean Average Precision (mAP) and the Cumulative Matching Characteristics (CMC). For CMC we report the rank-1, rank-5 and rank-10 accuracies. We evaluate the PAR's performance based on the label mean accuracy (ma), and the instance Precision, Recall and F1 scores.   

\subsection{Implementation Details}
Our method is designed for any existing MTL architecture. Therefore, we experiment with a very basic network architecture: a ResNet-18 \cite{he2016deep} backbone and two task-specific heads composed of a $1\times 1$ convolution, two fully-connected layers and a final classifier. The network outputs a feature vector and a classification vector for each task. Pre-trained Imagenet \cite{deng2009imagenet} weights are used as a starting point to train the MTL network. We train our model $\mathcal{F}(.)$ for $100$ epochs with batch-size $64$, using the Adam \cite{kingma2014adam} optimizer with the initial learning rate $lr=0.0003$ and a cosine learning rate scheduler. $\lambda = 1.0$ from Equation \ref{eq: tot loss}. 

We use an ID cross-entropy (CE) loss and an attribute binary cross-entropy (BCE) loss for the supervised reID ($\mathcal{L}^{GT}_1$) and PAR losses ($\mathcal{L}^{GT}_2$) respectively. For the reID consistency loss $\mathcal{L}^{semi}_1$, we compute a triplet loss on the reID output feature vectors, and a BCE loss on the dataset prediction output (part of the reID's classification output). For the PAR consistency loss $\mathcal{L}^{semi}_2$, we compute a DICE loss \cite{milletari2016v} on the PAR classification outputs and a triplet loss on the PAR output feature vectors. 

\subsection{Baselines}
We define five baselines to compare our KA against.

\noindent
\textbf{\textit{reID only}:} Model trained for reID with only reID data in a fully-supervised manner using $\mathcal{L}^{GT}_1$.

\noindent
\textbf{\textit{PAR only}:} Model trained for PAR with only PAR data in a fully-supervised manner using $\mathcal{L}^{GT}_2$.

\noindent
\textbf{\textit{reID ssl}:} Model trained for reID with reID and unlabeled data in a semi-supervised manner using $\mathcal{L}^{GT}_1$ and $\mathcal{L}^{semi}_1$.

\noindent
\textbf{\textit{PAR ssl}:} Model trained for PAR with PAR and unlabeled data in a semi-supervised manner using $\mathcal{L}^{GT}_2$ and $\mathcal{L}^{semi}_2$.

\noindent
\textbf{\textit{MTL baseline}:} Model trained for reID and PAR in a fully-supervised manner using only $\mathcal{L}^{GT}_1$ and $\mathcal{L}^{GT}_2$.

\subsection{Results} \label{subsec: res}
Our main quantitative results are shown in Table \ref{tab:res same domain}. The reID's mAP improves by $1.8\%$ points when trained alongside PAR (\textit{MTL baseline}). Its accuracy further improves by $0.4\%$ points when using our KA method of combining MTL and SSL to leverage unlabeled data. However, the performance drops for our \textit{reID ssl} baseline. We conjecture that this is due to different domain and unbalanced data. The PAR's F1 score improves by $0.8\%$ points when training with additional unlabeled data (\textit{PAR ssl} baseline). Its score further improves by $0.1\%$ points when using our KA method of combining SSL with MTL. In summary, our KA method outperforms all baselines for most metrics, meaning that training related tasks jointly and leveraging available but unlabeled data improves all tasks' performances. The qualitative results shown in Figure \ref{fig:qualita} further support this statement. Indeed, we notice that our KA method correctly predicts attributes (Female (F), Adult (A)) and re-identifies the person, unlike the single-task methods.

In Table \ref{tab:ablation} we outline our ablation study results. Firstly we show that using model augmentation instead of image augmentation \cite{sohn2020fixmatch} as a SSL framework yields higher performance for both tasks. Moreover, the network reaches its highest performance (mAP $=56.8\%$ for reID and F1 $=83.2\%$ for PAR) when computing an additional triplet loss on feature vectors for pseudo-supervision (\textit{KA w/ NetAug+Tri}), instead of only calculating a standard classification loss (\textit{KA w/ NetAug}). 

\begin{table}[h!]
\centering
\caption{\textbf{Ablation Study Results on ResNet-18.} We test different augmentation strategies and losses for the semi-supervised loss computation. Best results are shown in \textbf{bold}.}
\label{tab:ablation}
\resizebox{0.9\columnwidth}{!}{%
\begin{tabular}{l|cc|cc}
\hline
\textit{ResNet-18} & \multicolumn{2}{c|}{Market1501 (reID)} & \multicolumn{2}{c}{PA100K (PA)} \\ \cline{2-5} 
& \multicolumn{1}{c}{mAP (\%)} & \multicolumn{1}{c|}{rank-1 (\%)} & \multicolumn{1}{c}{ma (\%)} & \multicolumn{1}{c}{F1 (\%)} \\ \hline
KA w/ ImgAug     & 54.1      &  73.3      & 73.8      & 81.9      \\
KA w/ NetAug     & 55.8      &  74.5      & 74.2      & 82.7      \\
KA w/ NetAug+Tri & \bf{56.8} & \bf{75.9 } & \bf{74.4} & \bf{83.2} \\ \hline
\end{tabular}%
}
\end{table}

\section{Conclusion}
\label{sec:conclu}
In this work we focused on training multiple tasks for real-world applications, where datasets are labeled for only some of the tasks. Our method, KA is capable of utilizing all the available data by leveraging semi-supervised consistency regularization from model perturbation. The results show that learning multiple tasks with labeled and unlabeled data using KA leads to better performances than training networks for single tasks in a fully supervised manner. Future work would explore training different and larger combinations of tasks and datasets.


{\small
\bibliographystyle{ieee_fullname}
\bibliography{egbib}

\begin{thebibliography}{10}\itemsep=-1pt

\bibitem{chen2021semi}
Xiaokang Chen, Yuhui Yuan, Gang Zeng, and Jingdong Wang.
\newblock Semi-supervised semantic segmentation with cross pseudo supervision.
\newblock In {\em Conference on Computer Vision and Pattern Recognition
  (CVPR)}, 2021.

\bibitem{cordts2016city}
Marius Cordts, Mohamed Omran, Sebastian Ramos, Timo Rehfeld, Markus Enzweiler,
  Rodrigo Benenson, Uwe Franke, Stefan Roth, and Bernt Schiele.
\newblock The cityscapes dataset for semantic urban scene understanding.
\newblock In {\em Conference on Computer Vision and Pattern Recognition
  (CVPR)}, 2016.

\bibitem{deng2009imagenet}
Jia Deng, Wei Dong, Richard Socher, Li-Jia Li, Kai Li, and Li Fei-Fei.
\newblock Imagenet: A large-scale hierarchical image database.
\newblock In {\em Conference on Computer Vision and Pattern Recognition
  (CVPR)}. IEEE, 2009.

\bibitem{dhar2022eyepad++}
Prithviraj Dhar, Amit Kumar, Kirsten Kaplan, Khushi Gupta, Rakesh Ranjan, and
  Rama Chellappa.
\newblock Eyepad++: A distillation-based approach for joint eye authentication
  and presentation attack detection using periocular images.
\newblock In {\em Conference on Computer Vision and Pattern Recognition
  (CVPR)}, 2022.

\bibitem{gu2022contrastive}
Ran Gu, Jingyang Zhang, Guotai Wang, Wenhui Lei, Tao Song, Xiaofan Zhang, Kang
  Li, and Shaoting Zhang.
\newblock Contrastive semi-supervised learning for domain adaptive segmentation
  across similar anatomical structures.
\newblock {\em IEEE Transactions on Medical Imaging}, 42, 2022.

\bibitem{he2016deep}
Kaiming He, Xiangyu Zhang, Shaoqing Ren, and Jian Sun.
\newblock Deep residual learning for image recognition.
\newblock In {\em Conference on Computer Vision and Pattern Recognition
  (CVPR)}, 2016.

\bibitem{hong2020beyond}
Yan Hong, Li Niu, Jianfu Zhang, and Liqing Zhang.
\newblock Beyond without forgetting: Multi-task learning for classification
  with disjoint datasets.
\newblock In {\em International Conference on Multimedia and Expo (ICME)}.
  IEEE, 2020.

\bibitem{huang2020partly}
Chao Huang, Hui Tang, Wei Fan, Yuan Xiao, Dingjun Hao, Zhen Qian, Demetri
  Terzopoulos, et~al.
\newblock Partly supervised multi-task learning.
\newblock In {\em International Conference on Machine Learning and Applications
  (ICMLA)}. IEEE, 2020.

\bibitem{huang2021age}
Zhizhong Huang, Junping Zhang, and Hongming Shan.
\newblock When age-invariant face recognition meets face age synthesis: A
  multi-task learning framework.
\newblock In {\em Conference on Computer Vision and Pattern Recognition
  (CVPR)}, 2021.

\bibitem{kendall2018multi}
Alex Kendall, Yarin Gal, and Roberto Cipolla.
\newblock Multi-task learning using uncertainty to weigh losses for scene
  geometry and semantics.
\newblock In {\em Proceedings of the IEEE Conference on Computer Vision and
  Pattern Recognition (CVPR)}, 2018.

\bibitem{kim2018disjoint}
Dong-Jin Kim, Jinsoo Choi, Tae-Hyun Oh, Youngjin Yoon, and In~So Kweon.
\newblock Disjoint multi-task learning between heterogeneous human-centric
  tasks.
\newblock In {\em Winter Conference on Applications of Computer Vision (WACV)}.
  IEEE, 2018.

\bibitem{kingma2014adam}
Diederik~P Kingma and Jimmy Ba.
\newblock Adam: A method for stochastic optimization.
\newblock {\em arXiv preprint arXiv:1412.6980}, 2014.

\bibitem{kokkinos2017ubernet}
Iasonas Kokkinos.
\newblock Ubernet: Training a universal convolutional neural network for low-,
  mid-, and high-level vision using diverse datasets and limited memory.
\newblock In {\em Conference on Computer Vision and Pattern Recognition
  (CVPR)}, 2017.

\bibitem{lin2019improving}
Yutian Lin, Liang Zheng, Zhedong Zheng, Yu Wu, Zhilan Hu, Chenggang Yan, and Yi
  Yang.
\newblock Improving person re-identification by attribute and identity
  learning.
\newblock {\em Pattern recognition}, 95, 2019.

\bibitem{liu2017hydraplus}
Xihui Liu, Haiyu Zhao, Maoqing Tian, Lu Sheng, Jing Shao, Junjie Yan, and
  Xiaogang Wang.
\newblock Hydraplus-net: Attentive deep features for pedestrian analysis.
\newblock In {\em International Conference on Computer Vision (ICCV)}, 2017.

\bibitem{liu2015deep}
Ziwei Liu, Ping Luo, Xiaogang Wang, and Xiaoou Tang.
\newblock Deep learning face attributes in the wild.
\newblock In {\em International Conference on Computer Vision (ICCV)}, 2015.

\bibitem{milletari2016v}
Fausto Milletari, Nassir Navab, and Seyed-Ahmad Ahmadi.
\newblock V-net: Fully convolutional neural networks for volumetric medical
  image segmentation.
\newblock In {\em Iinternational Conference on 3D Vision (3DV)}. IEEE, 2016.

\bibitem{ouali2020overview}
Yassine Ouali, C{\'e}line Hudelot, and Myriam Tami.
\newblock An overview of deep semi-supervised learning.
\newblock {\em arXiv preprint arXiv:2006.05278}, 2020.

\bibitem{ruder2017overview}
Sebastian Ruder.
\newblock An overview of multi-task learning in deep neural networks.
\newblock {\em arXiv preprint arXiv:1706.05098}, 2017.

\bibitem{schumann2017person}
Arne Schumann and Rainer Stiefelhagen.
\newblock Person re-identification by deep learning attribute-complementary
  information.
\newblock In {\em Conference on Computer Vision and Pattern Recognition
  Workshops (CVPRW)}, 2017.

\bibitem{sohn2020fixmatch}
Kihyuk Sohn, David Berthelot, Nicholas Carlini, Zizhao Zhang, Han Zhang,
  Colin~A Raffel, Ekin~Dogus Cubuk, Alexey Kurakin, and Chun-Liang Li.
\newblock Fixmatch: Simplifying semi-supervised learning with consistency and
  confidence.
\newblock {\em Advances in neural information processing systems}, 33, 2020.

\bibitem{specker2023upar}
Andreas Specker, Mickael Cormier, and J{\"u}rgen Beyerer.
\newblock Upar: Unified pedestrian attribute recognition and person retrieval.
\newblock In {\em Winter Conference on Applications of Computer Vision (WACV)},
  2023.

\bibitem{vandenhende2021multi}
Simon Vandenhende, Stamatios Georgoulis, Wouter Van~Gansbeke, Marc Proesmans,
  Dengxin Dai, and Luc Van~Gool.
\newblock Multi-task learning for dense prediction tasks: A survey.
\newblock {\em IEEE transactions on pattern analysis and machine intelligence},
  44, 2021.

\bibitem{wang2022semi}
Yufeng Wang, Yi-Hsuan Tsai, Wei-Chih Hung, Wenrui Ding, Shuo Liu, and
  Ming-Hsuan Yang.
\newblock Semi-supervised multi-task learning for semantics and depth.
\newblock In {\em Winter Conference on Applications of Computer Vision (WACV)},
  2022.

\bibitem{xie2020unsupervised}
Qizhe Xie, Zihang Dai, Eduard Hovy, Thang Luong, and Quoc Le.
\newblock Unsupervised data augmentation for consistency training.
\newblock {\em Advances in neural information processing systems}, 33, 2020.

\bibitem{zhang2021survey}
Yu Zhang and Qiang Yang.
\newblock A survey on multi-task learning.
\newblock {\em IEEE Transactions on Knowledge and Data Engineering}, 34, 2021.

\bibitem{zhen2020joint}
Mingmin Zhen, Jinglu Wang, Lei Zhou, Shiwei Li, Tianwei Shen, Jiaxiang Shang,
  Tian Fang, and Long Quan.
\newblock Joint semantic segmentation and boundary detection using iterative
  pyramid contexts.
\newblock In {\em Conference on Computer Vision and Pattern Recognition(CVPR)},
  2020.

\bibitem{zheng2015scalable}
Liang Zheng, Liyue Shen, Lu Tian, Shengjin Wang, Jingdong Wang, and Qi Tian.
\newblock Scalable person re-identification: A benchmark.
\newblock In {\em International Conference on Computer Vision (ICCV)}, 2015.

\bibitem{zhu2005semi}
Xiaojin~Jerry Zhu.
\newblock Semi-supervised learning literature survey.
\newblock 2005.

\end{thebibliography}
}

\end{document}